\pdfoutput=1
\documentclass[11pt]{article}

\usepackage[final]{acl}

\usepackage{times}
\usepackage{latexsym}

\usepackage[T1]{fontenc}
\usepackage[utf8]{inputenc}

\usepackage{microtype}

\usepackage{inconsolata}

\usepackage{enumerate}
\usepackage{amsthm}
\usepackage{graphicx}
\usepackage{epstopdf}
\usepackage{caption}
\usepackage{color}
\usepackage{enumitem}
\usepackage{array}
\usepackage{tabularx}
\usepackage{multirow}
\usepackage{bm}
\usepackage{booktabs}

\usepackage{caption}
\usepackage{subcaption}
\usepackage{listings}
\usepackage{wrapfig}
\usepackage{bbding}
\usepackage{colortbl}
\usepackage{fontawesome}

\usepackage[dvipsnames]{xcolor}

\usepackage{authblk}
\usepackage{tcolorbox}
\tcbuselibrary{breakable}
\tcbuselibrary{skins}

\usepackage{float}

\usepackage{etoolbox}
\preto{\abstractkeywords}{\nolinenumbers} 

\newcommand{\faCheckG}{\textcolor{ForestGreen}{\faCheck}}
\newcommand{\faTimesR}{\textcolor{Red}{\faTimes}}

\title{Long-context Language Models Fail in Basic Retrieval Tasks Without Sufficient Reasoning Steps}

\author[a,c]{Yijiong Yu}

\author[a]{\\Yongfeng Huang}
\author[a]{Zhixiao Qi}

\author[c]{\\Wei Wang}
\author[c]{Weifeng Liu}
\author[c]{Ran Chen}
\author[c]{Ji Pei}

\affil[a]{Tsinghua University}
\affil[c]{OpenCSG}
\begin{document}

\maketitle

\begin{abstract}
Long-context language models (LCLMs), characterized by their extensive context window, are becoming popular. However, despite the fact that they are nearly perfect at standard long-context retrieval tasks, our evaluations demonstrate they fail in some basic cases. Later, we find they can be well addressed with a sufficient number of reasoning steps, guided by specific CoT prompts. This result emphasizes the potential necessity of solving specific long-context tasks using long-CoT methods, while previous long-context benchmarks always ignore the necessity of long reasoning for long-context tasks and treat them as direct QA tasks. Our code and datasets are available at \url{https://github.com/yuyijiong/hard_retrieval_for_llm}.

%Thus we propose a critical viewpoint that there are currently no perfect solutions for current LCLMs to solve all types of retrieval tasks.

%Our work reveals some novel properties of retrieval tasks and LCLMs,

% This finding reminds the developers and users of LCLMs that relying on LCLMs to directly perform even basic retrieval tasks may be unreliable, rather, a sufficiently long reasoning process is necessary. 

\end{abstract}

\section{Introduction}
In the past years, long-context language models (LCLMs) such as GPT-4o-128k \citep{} and Gemini-1.5-1000k \citep{gemini_team_gemini_2024} have surged in popularity, raising questions about their efficacy in handling extended context tasks. While various LCLMs have demonstrated perfect long-context retrieval ability by passing the ``Needle in a Haystack'' test \citep{gkamradt_llmtest_needleinahaystack_2023} in over 100k context length, benchmarks like Loogle \citep{li_loogle_2023}, Ruler \citep{hsieh_ruler_2024}, and Loong \citep{wang_leave_2024} have highlighted their shortcomings in more complex tasks. 

However, previous long-context benchmarks, despite the variety of task types, typically categorized tasks based on their intuitive forms rather than their inherent difficulty or nature, making it hard to rank their true difficulties or determine which sub-task is the bottleneck. What is more, they usually ignore the necessity of reasoning, and simply regard the long-context task as a direct QA task, as they always set the prompt as ``directly give a concise answer''. This leads to an unclear ability boundary of LCLMs and difficulties of long-context tasks.

Some recent works \cite{li_needlebench_2024,wang2024mathhayautomatedbenchmarklongcontext,fei2024retrievalmeetsreasoningdynamic} have tested models on challenging long-context reasoning tasks which necessitate CoT, such as multi-query, multi-hop questions, or questions requiring logic. However, these tasks only increase the complexity of the question, in other words, the number of reasoning steps are just determined by the question itself (so we call these tasks multi-step-question), but not the context (a long context task usually consists of a context and a question). As a result, they still do not find long-context tasks with a straightforward and short question may also necessitate long-CoT. 

%Neither, there are few works exploring the importance of CoT reasoning for long-context retrieval tasks. 

\begin{figure*}[htbp]
    \centering
\subfloat[]{
\label{fig:example task}
    \includegraphics[width=0.6\linewidth]{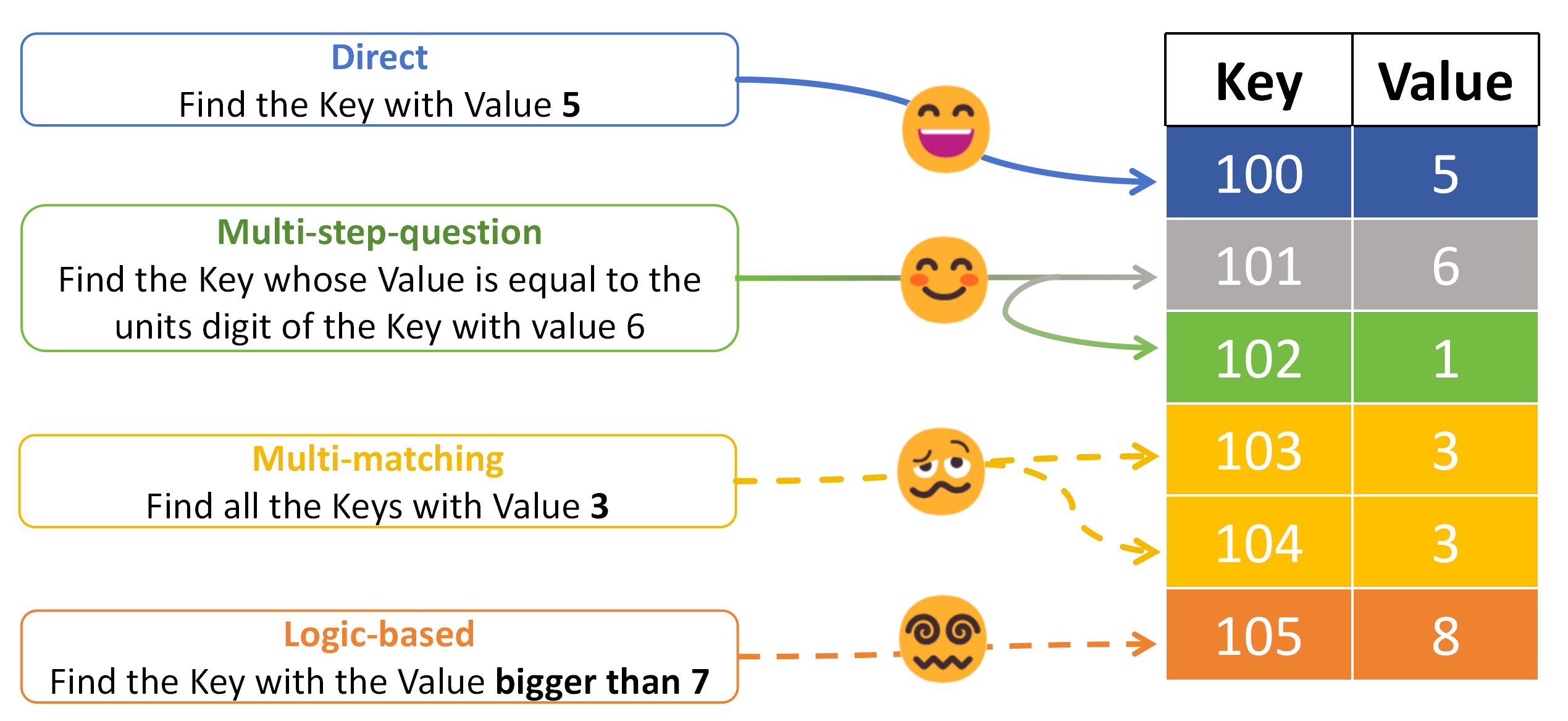}}
\subfloat[]{
\label{fig:score 4 task}
    \includegraphics[width=0.3\linewidth]{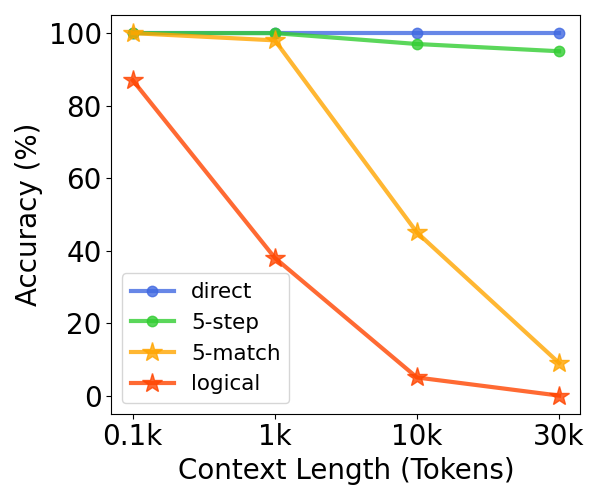}}
    \caption{(a) Examples of direct, multi-step-question, multi-matching and logic-based Key-Value retrieval. (b) Accuracy of GPT-4o in different KV retrieval tasks, as the context length increases. 5-step means a 5-hop-question retrieval task. 5-match means a multi-matching retrieval task with 5 matching items for one query. Logical means logic-base retrieval.}
    
\end{figure*}

% \begin{figure}[htbp]
%     \centering
    
% \includegraphics[width=0.5\linewidth]{figures/gpt4_4_perform.png}

%     \caption{Accuracy of GPT-4o in different KV retrieval tasks, as the context length increases. 5-step means a multi-step-question retrieval task requiring at least 5 times of retrieval. 5-match means a multi-matching retrieval task with 5 matching items for one query. Logical means logic-base retrieval.}
%     \label{fig:score 4 task}
% \end{figure}

% Neither, they did not explore the importance of reasoning for long-context retrieval. This has raised our suspicion: are LCLMs really good at all types of retrieval tasks, even without CoT?

Therefore, in order to more scientifically define the difficulty of long-context tasks, we conduct a detailed analysis (detailed in Appendix \ref{app:task belong} and \ref{app:detail each task}) of various tasks (especially retrieval-based tasks) from previous long-context benchmarks. We summarize 2 basic cases, which exhibit much higher difficulty and exhibit distinct properties for LCLMs: multi-matching retrieval and logic-based retrieval. Multi-matching retrieval involves recalling multiple items simultaneously, and logic-based retrieval involves logical judgment within retrieval criteria. Although they are both ``basic'' retrieval problems in a straightforward form, 
our experiments, as shown in Figure \ref{fig:score 4 task}, demonstrate they can exceed the ability boundary of LCLMs as the context length grows in non-CoT settings, manifested as a rapid drop in accuracy when the context length exceeds a certain value. In contrast, the performance of traditional ones (direct retrieval and multi-step-question retrieval, which are previously researched) is stable. Additionally, we conduct some analytic experiments on the micro-level behaviors of LCLMs to explain why they always fail in these cases.

Later, we test whether CoT prompting\cite{wei_chain--thought_2023} could help in these cases. At first, we try to simply asking the model to think step-by-step \cite{wei_chain--thought_2023}, but find it only makes the model generate unnecessary steps and hardly improves retrieval accuracy. But when trying more well-designed prompts, we identify a sufficient condition of resolving these problems: adopting sufficient reasoning steps. Specifically, ``sufficient'' means the number is at least $N+\frac{n^2+n}{2}$, where $N$ signify total items in the context and $n$ signify the number of items to retrieve. 

Based on the above findings, we categorize long-context retrieval problems into three basic types: 
(1) Simple: they do not need CoT (mainly simple retrieval)  
(2) Question-Difficult: they need CoT, usually with a few steps, the number of steps are determined by the question itself (mainly multi-step-question retrieval) 
(3) Context-Difficult: they also need CoT, but usually with numerous steps, and the number of steps are determined by the context, specifically, the length of the context or the number of items in the context (such as multi-matching and logic-based retrieval)

In order to help readers distinguish the difference between Question-Difficult and Context-Difficult tasks more clearly, in Figure \ref{fig:example task} and Appendix \ref{app:distinguish}, we use examples to detail the formal distinctions between multi-matching retrieval, logic-based retrieval and their question-difficult variants (multi-query and logic-required retrieval). 

In conclusion, certain types of tasks are actually beyond the traditional ``long-context capability'' of LCLMs, but rather require a long reasoning process with sufficient steps. So, we summarize two helpful insights for the long-context research: (1) to solve a wider range of long context problems, relying solely on a larger context window is not enough, and long reasoning may also be needed; (2) for a very long context, long reasoning requires too many steps, making it inefficient, so it is still important to find some novel ways to handle long texts efficiently.

% We are the first to classify retrieval tasks from the perspective of reasoning instead of traditional "long-context capability", and emphasize the importance of 'long CoT' in long context tasks. The evidence is that we find some certain types of retrieval tasks are always unsolvable even for powerful long-context LMs, unless adopting long-CoT.  While previous benchmarks always ignore the necessity of reasoning for long-context tasks, even though they include tasks requiring reasoning. 

% We hope this critical finding can guide the developers to perceive various types of long-context problems from a more scientific and intrinsic perspective, and utilize LCLMs in a more appropriate way.

% Although our experiments only test on simple synthetic samples, the conclusion can also provide fundamental insights into why LCLMs are still bad at some high-level, real-world long-context tasks.

%实验1
\section{LCLMs Perform Poorly in Multi-matching and Logic-based Retrieval}

In this section, we conduct evaluations on 5 popular long-context models with context windows over 128k tokens, including Phi-3.5-mini-instruct (phi-3.5) \citep{abdin_phi-3_2024}, Meta-Llama-3.1-70B-Instruct (llama3.1-70b) \citep{dubey_llama_2024}, Deepseek-V2.5 (deepseek) \citep{deepseek-ai_deepseek-v2_2024}, Gemini-1.5-flash (gemini) \citep{gemini_team_gemini_2024}, and GPT-4o-2024-08-06 (gpt-4o) \citep{openai2024gpt4ocard}. We find that, multi-matching and logic-based retrieval are really hard, nearly unsolvable for current LCLMs in long context. In contrast, multi-step-question retrieval proves to be much easier. 

In the main content we only show the results of gemini and gpt-4o in multi-matching and logic-based retrieval. The complete evaluation results of other models and direct retrieval or multi-step-question retrieval are in Appendix \ref{full result}.

\begin{table*}[tb]
\centering
%\resizebox{\columnwidth}{!}{
\begin{tabular}{lc|ccc|cc}
\toprule
\multirow{2}{*}{Model} & \multirow{2}{*}{Num Matches ($n$)} & \multicolumn{3}{c|}{KV Retrieval} & \multicolumn{2}{c}{Student Resume Retrieval} \\
\cmidrule{3-7}
\multicolumn{2}{c|}{} &$N$=10 &$N$=100 &$N$=1000 &$N$=10 &$N$=100 \\
\midrule
\multirow{4}{*}{gemini} 
& 1 & 100 & 100 & 94 & 100 & 96 \\
& 5 & 100 & 82 & 66 & 100 & 36 \\
& 10 & / & 48 & 18 & / & 8 \\
& 20 & / & 35 & 2 & / & 1 \\
\midrule
\multirow{4}{*}{gpt-4o} 
& 1 & 100 & 100 & 60 & 100 & 100 \\
& 5 & 100 & 98 & 45 & 100 & 65 \\
& 10 & / & 85 & 5 & / & 0 \\
& 20 & / & 50 & 0 & / & 0 \\
\bottomrule
\end{tabular}
%}
\caption{Accuracy (\%) of two tasks: KV Retrieval and Student Resume Retrieval, requiring retrieving all matches under varying $N$.}
\label{tab: combined score multi short}
\end{table*}

\begin{table*}[tb]
\centering
%\resizebox{\columnwidth}{!}{
\begin{tabular}{l|cccc|ccc}
\toprule
\multirow{2}{*}{Model} & \multicolumn{4}{c|}{KV Retrieval} & \multicolumn{3}{c}{Student Resume Retrieval} \\
\cmidrule{2-8}
 & $N$=4~~ & $N$=10~ & $N$=100& $N$=1000 & $N$=4~~  & $N$=10~ & $N$=100 \\
\midrule
gemini & 78 & 33 & 6 & 0 & 92 & 80 & 12 \\
gpt-4o & 97 & 87 & 38 & 5 & 100 & 92 & 30 \\
\bottomrule
\end{tabular}
%}
\caption{The accuracy (\%) on logic-based KV retrieval and Student Resume Retrieval.}
\label{tab:combined logic short}
\end{table*}

\subsection{Experimental Settings for Evaluation}
\label{sec:data}

We create two fully synthetic datasets: Key-Value Pair Retrieval and Student Resume Retrieval. We use $N$ to denote the total number of items in the input context, and $n$ to denote the number of items to be retrieved and outputted in the model's response. We must point out that our problems are simplified, only used to reflect the abilities of LCLMs, but not real-world tasks, as it would be easy to solve using code such as SQL query.

In Key-Value pair retrieval, the context is a JSON-formatted dictionary consisting of $N$ randomly generated Key-Value pairs. The Key is a 10-digit string, and the Value is a positive integer. The question is appended to the context and varies based on the task type. For multi-matching, the model must retrieve all Keys associated with a given Value. For logic-based retrieval, the model needs to identify the Key with the Value within a specified range.

% \begin{tcolorbox}[colback=white, title=Context (KV Retrieval), colbacktitle=gray,breakable]
% \small{{\sffamily JSON data with 3000 Key-Value pairs:\\
% \{"1532968704": 78, ``5921306748": 84, ``3742815096": 47, ......, ``3276918540": 76\}}}
% \end{tcolorbox}

% \begin{tcolorbox}[colback=white, title=Question (KV Retrieval), colbacktitle=blue!40!white,breakable]
% \small{{\sffamily
% \textbf{Direct retrieval:} In the above JSON data, please find the value of the key '6978024153'. Provide your final answer in the format ``value: \{answer\}". 
% \tcbline
% \textbf{Multi-matching retrieval:} In the above JSON data, please find all the keys with the value 0. Provide your answer (keys separated by commas) in the format ``keys: \{answer\}".
% \tcbline
% \textbf{Logic-based retrieval:} In the above JSON data, please find the key (only one) whose value is greater than 223 and smaller than 278. Provide your answer in the format ``key: \{answer\}".}}
% \end{tcolorbox}

In the Student Resume Retrieval task, all original resumes are fictional and generated by GPT-4o. The context includes $N$ rows, each detailing a fictional college student's information, such as name, age, graduation school, interests, GPA, and a short self-introduction. In multi-matching problems, the task is to retrieve all students graduating from a specified university. For logic-based retrieval, the problem is to identify the student whose GPA falls within a specified range. All GPAs are rounded to two decimal places and range from 0 to 5. Moreover, to test more types of logic, we design another logic-based retrieval task, where the problem is to identify the student whose interest belongs to a certain field.

Examples of the prompts of KV retrieval or student resume retrieval, and more information on how we construct samples and why we choose such formats is shown in the Appendix \ref{app:data}.

In all experiments, we set the temperature to 0, and max generated tokens to 512. We use accuracy as the metric to ensure strictness. For each setting, we use 100 test samples.

\subsection{Model Performance on Multi-matching Retrieval}

The results presented in Table \ref{tab: combined score multi short} reveal that when only 1 matching item is present, larger models, such as Gemini \citep{gemini_team_gemini_2024}, demonstrate superior performance, achieving an accuracy of up to 94\% even in lengthy contexts. However, with the introduction of multiple matching items, such as 5 or 10, the accuracy of all language models rapidly declines to nearly zero, particularly evident in the more realistic scenario of Student Resume Retrieval. This trend suggests that the inherent difficulty of the task is consistently challenging across models of varying sizes.

We also count different types of errors and detail the ratio of over-selection, under-selection and mis-selection. The complete results are in Appendix \ref{app:detail multi match}.

\subsection{Model Performance on Logic-based Retrieval}
\label{logic score}

As illustrated in Table \ref{tab:combined logic short}, in logic-based retrieval, when the context contains only 4 options, LLMs successfully select the correct item in most cases, indicating that these models possess logical judgment capabilities. However, for both datasets, all tested models struggle with these tasks as the context length increases, consistently retrieving an incorrect item whose value lies outside the specified range. The results of other models and the classification-based logical retrieval shown in Appendix \ref{full result} exhibit the same trend. 

% So it is reasonable to anticipate that these difficulties will be further exacerbated in more complex scenarios that require logical judgment beyond mere numerical comparison.

%Similarly, the use of CoT \citep{wei_chain--thought_2023} prompts does not mitigate this issue for the same reasons.

\subsection{Explore the Cause of Failure}
To better understand why powerful LLMs perform so badly, we analyze their behaviors on multi-matching and logic-based retrieval tasks (when CoT prompt is not used) by more in-depth experiments based on probing (similar to \citealt{lu2024insightsllmlongcontextfailures}), and get some interesting findings. Specifically, we probe whether the model has already obtained the information of the gold KV in KV retrieval task in every layer, from the hidden states of the anchor token (the token right before the retrieval answer). More details and analysis about this experiment are shown in Appendix \ref{analytical}. These findings further prove these problems cannot be easily solved only relying on a large context window.

\begin{figure}[tbp]
    \centering
    
\includegraphics[width=0.8\linewidth]{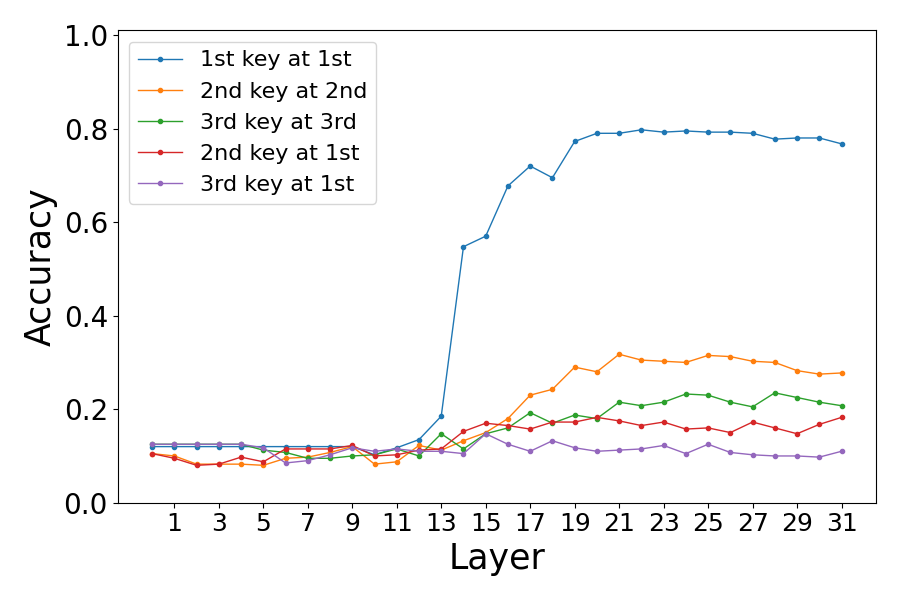}

    \caption{The accuracy of probing from hidden states of each layer of different anchor tokens when predicting the first, second, and third Key. For example, ``2nd key at 1st'' means we probe from the hidden states of the 1st anchor token but use the 2nd anchor token (the 2nd gold Key's first digit) as the label for training and testing.}
    \label{fig: 3 matching probe main}
\end{figure}

For multi-matching retrieval, the result is in Figure \ref{fig: 3 matching probe main}. The example is 3-matching retrieval, so there are 3 anchor tokens. The curve of ``1st key at 1st'' is high after layer 14, indicating retrieving the first item is easy. However, the accuracy of probing the 2nd or 3rd item from the 1st anchor is always low, which means the model has to retrieve one by one but not all at once. Moreover, the accuracy of probing the 1st, 2nd or 3rd item respectively from the 1st, 2nd or 3rd anchor is greatly decreasing in sequence, which means the difficulty of retrieving later items in multi-matching retrieval is gradually increasing. Similar trends have also been discovered in FACT \cite{wang_fact_2024}. This may reveal why LLMs always fail to retrieval all the matching items.

\begin{figure}[tbp]
    \centering
    
\includegraphics[width=1\linewidth]{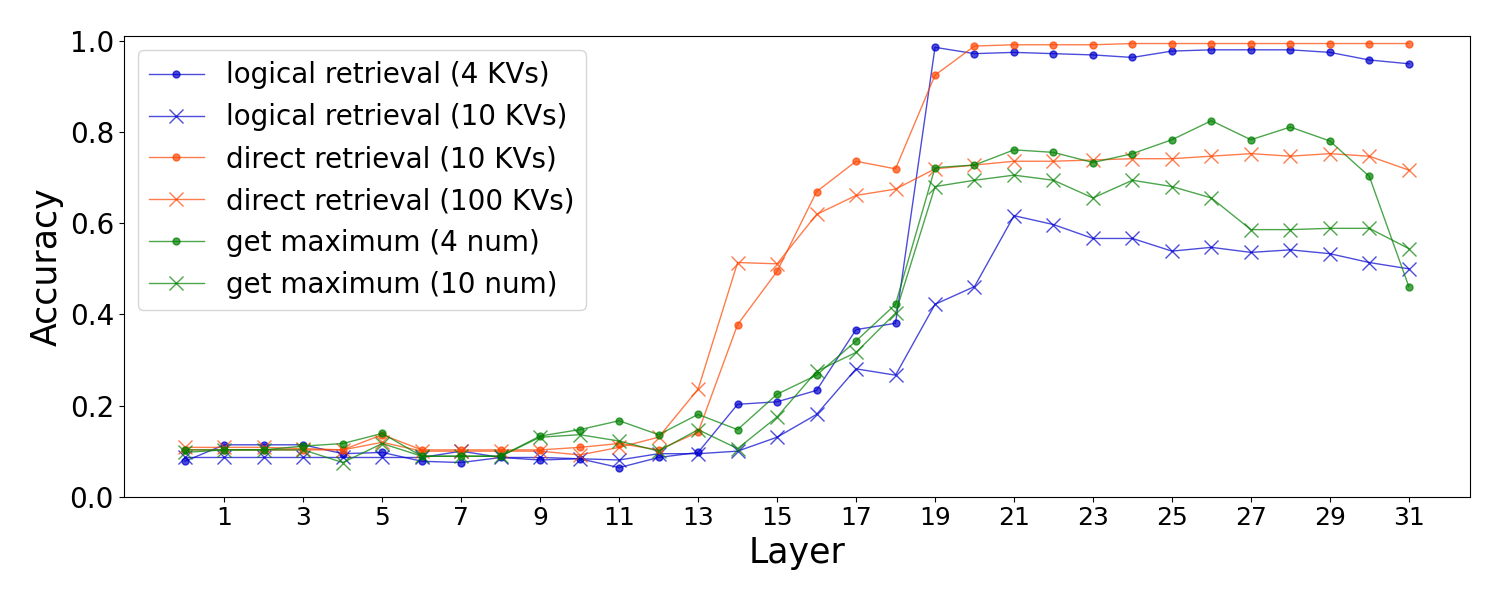}

    \caption{Linear probing accuracy of each layer of 3 tasks: direct KV retrieval, logical KV retrieval, and an arithmetic problem (getting the maximum value among N numbers).}
    \label{fig:logic probing main}
\end{figure}

For logic-based retrieval, the result is in Figure \ref{fig:logic probing main}. We use a traditional direct retrieval task, and a simple arithmetic task as comparisons. We can clearly see that the point where the curve starts to quickly rise is similar for logic-based retrieval and the arithmetic problem, but that of direct retrieval is much earlier. This indicates the difficulty of logic-based retrieval may resemble a math problem involving many steps of logic judgments or arithmetic calculations, which cannot be performed by LLMs simultaneously without an explicit CoT process.

% Mainly, we find: (1) the difficulty of retrieving each item one-by-one in multi-matching retrieval is gradually increasing, and (2) the difficulty of logic-based retrieval may stem from LLMs' inability to perform multiple logic judgments or arithmetic calculations simultaneously without outputting extra tokens. 

\begin{table*}[tb]
\centering
%\resizebox{\columnwidth}{!}{

\begin{tabular}{l|cc|cc|c}
\toprule
\multirow{2}{*}{Prompt} & \multicolumn{2}{c|}{KV Retrieval} & \multicolumn{2}{c|}{Student Resume Retrieval} & \multirow{2}{*}{Output Tokens} \\
\cmidrule{2-5}
 & logic-based & multi-matching & logic-based & multi-matching & \\
 \midrule
standard       &  38    & 50  &  30  & 0     & 12        \\
step-by-step   &  47    & 52  &  37  & 10    & 340       \\
one-by-one     &  100   & 90  &  100 & 20    & 1500       \\
add to list     &  100   & 92  &  100 & 90    & 1700       \\

\bottomrule
\end{tabular}
%}
\caption{Performance and output tokens (reflecting the number of reasoning steps) of GPT-4o with 4 types of prompts. In all the experiments, $N$ is set to 100. In multi-matching retrieval, $n$ is set to 20 in KV retrieval and 10 in Student Resume Retrieval. In logic-based retrieval, $n$ is always 1.}
\label{tab:one-by-one}
\end{table*}

\section{Using CoT With Sufficient Steps}
Since asking LLMs to directly do retrieval within one step leads to bad performance, it is natural to wonder if CoT-like prompt might help. Thus, we design special prompts to guide step-by-step reasoning, to test if this could help model solve these retrieval tasks. In experiments, we employ 4 types of prompts: 

(1) Standard prompt: the default prompt without CoT, serves as the baseline.

(2) Step-by-step prompt: this is the traditional CoT prompt which tells the model to ``think step by step, but do not check each item one by one''. With this prompt, LLMs usually automatically takes 3 or 4 steps. 

(3) One-by-one prompt: this prompt tells the model ``You should first examine every item one by one to give the judgement (yes/no) on whether it meets the requirement''. With this prompt, LLMs will at least takes $N$ steps to check each item.  

(4) One-by-one with adding to list prompt, examining every item one by one and add the correct one into the list. Besides $N$ steps for checking each item, it will take additional $\frac{n^2+n}{2}$ steps to maintain a gradually growing list.

As shown in Table \ref{tab:one-by-one}, we find the accuracy on logic-based retrieval is greatly improved to 100\% with one-by-one prompt, though it costs hundreds of times more time. In contrast, traditional CoT prompt with limited output length cannot improve much. For multi-matching retrieval, it can only be solved by the 4th prompt, which costs the most output tokens and the longest time. Examples of our prompts and the model's responses are shown in Appendix \ref{app:one by one}.

Therefore, we conclude that LLM can indeed solve these context-difficult problems if 
we scale the reasoning process to $N+\frac{n^2+n}{2}$ steps, with a well-designed prompt. However, this method has huge latency when the context is very long or there are too many items to retrieve. So far, we have not found a solution which achieves both accuracy and efficiency purely relying on LLM itself (some other technologies may help, shown in Appendix \ref{app:potential}), which indicates the original intention of LCLMs, to process massive information simultaneously, has not been fully achieved. However, we find the reasoning steps for these retrieval tasks are basically simple and parallel (not involving very complicated or interlocking logic), which indicates there may be methods to compress them to improve efficiency.

% indicates a deficiency of current LCLMs: the original intention of LCLMs is to process massive input information simultaneously, but this may be impossible, because outputting sufficient steps is needed to ensure an accurate retrieval, no matter how many items need to be retrieved in the input context.

% It is easy to anticipate that this difficulty will be even exacerbated in more complex scenarios where the logical judgement is more than numeric comparison. Therefore, we can assert with confidence that such problems are nearly unsolvable for current LLMs in long context.

\section{Conclusion}
In this paper, we use evaluation on synthetic datasets to demonstrate that LCLMs always struggle to solve some basic cases, such as multi-matching retrieval and logic-based retrieval. Then, we find they must be addressed by a sufficiently long reasoning process guided by specific prompts. Therefore, we highlight that merely extending the context window size and relying solely on the ``long-context capability'' of LCLMs cannot address all types of long-context tasks. On the other hand, although long reasoning process can improve accuracy, it is too time-consuming and inefficient. Thus, we urge more novel perspectives and approaches to more efficiently and accurately exploit long contexts.

% This paper examines the behavior of LLMs in various situations of long-context tasks to identify their limitations and find the problems' natures. We find current LCLMs may encounter difficulties even managing basic situations, such as multi-matching and logic-based retrieval,  Through our studies, we do not aim to propose more challenging benchmarks to drive further improvements and optimizations in LCLMs. Rather, we demonstrate a tough reality: while LCLMs are intended to process vast amounts of input data simultaneously, there exist certain long-context tasks which always remain unattainable for LCLMs to solve in one step. Therefore, future research should focus on addressing the challenges associated with numerous steps, rather than merely extending the context window of LLMs.

% Our study highlights that the method that solely relies on training a long-context LLM to directly generate answers for long-context tasks may encounter inherent limitations. More innovative approaches are still necessary for further exploiting long contexts.

\section{Limitations}
Our study only use synthetic retrieval tasks to reflect LCLMs' abilities in different settings. Although it is sufficient to illustrate the issue, evaluating in complex real-world scenarios would be better.

% We have summarized 2 basic cases making long-context retrieval hard, but there may also be other cases or factors.

Our proposed prompting method guiding the model to reason are only used to demonstrate the ability of LCLMs, rather than a solution for real-life problems. 

% The reasoning steps in our experiment are almost parallel and flat, which means it does not involve a complex and deep chain of reasoning. So, maybe there are some solutions to accelerate or compress such reasoning process.

%\clearpage
\bibliography{custom}
%\bibliographystyle{acl_natbib}
%\bibliography{acl_latex}

%\clearpage
\appendix

\section{Experimental Settings for Data Construction and Model Evaluation}
\label{app:data}

\subsection{Data Construction}
To prevent data contamination, we create two fully synthetic datasets: Key-Value Pair Retrieval and Student Resume Retrieval. These datasets make it easy for us to make controlled changes to the input context size and problem types.

In Key-Value pair retrieval, the context is a JSON-formatted dictionary consisting of $N$ randomly generated Key-Value pairs. The Key is a 10-digit string, and the Value is a positive integer. The question is appended to the context and varies based on the task type. For multi-matching, the model must retrieve all Keys associated with a given Value. For logic-based retrieval, the model needs to identify the Key with the Value within a specified range.

In the Student Resume Retrieval task, all original resumes are fictional and generated by GPT-4o. The context includes $N$ rows, each detailing a fictional college student's information, such as name, age, graduation school, interests, GPA, and a short self-introduction. In multi-matching problems, the task is to retrieve all students graduating from a specified university. For logic-based retrieval, the problem is to identify the student whose GPA falls within a specified range or interest belongs to a certain category. All GPAs are rounded to two decimal places and range from 0 to 5. 

We denote $N$ as the total number of items (an item corresponds to either a key-value pair or a student) in the input context and $n$ as the number of items to be retrieved. In KV retrieval, $N$ is set to 4, 10, 100, 1000 and the corresponding context length is 0.04k, 0.1k, 1k and 10k tokens respectively. In student resume retrieval, $N$ is set to 4, 10 and 100 and the corresponding context length is 0.3k, 0.6k and 6k respectively. For logic-based retrieval, $n$ is set to 1, and the model is informed that only one correct item exists. In multi-matching ($n$-matching) retrieval, $n$ is set to different values from 1 to 20, and the model is not informed of the number of matching items. The gold items (i.e. the items to be retrieved) are distributed at random positions within the context. For each problem setting, we construct 200 test samples with different context. 

For direct or logic-based retrieval where $n=1$, each question has the exact answer, making evaluation straightforward and objective. We use exact-match evaluation and accuracy as the metric for all tasks. For multi-matching retrieval, the model may generate a list of items as the answer, which may be partially correct and random in order. To calculate accuracy, we only consider the prediction correct if it is an totally exact match (without considering the order of the items) to the reference set, i.e., both over-selection and under-selection are considered incorrect. 
 
We have experimented with different context formats, such as declarative sentences, JSON dictionary, and Markdown tables, as well as placing the question before the context. These variations show no significant impact on performance. Moreover, although logic-based retrieval can be more advanced forms, we restrict it to basic forms such as numeric comparison for simplicity. Therefore, we standardize the format of all prompts to align with the examples provided above across all experiments.

% Note that, our datasets are just a simplistic version of tasks in real-life scenarios, designed to illustrate the limitations of LLM's capabilities. But the difficulty of these issues themselves is not high, because the context is actually structured, letting it easy to solve by traditional methods like template matching or SQL searching.

\subsection{Data Example}
\label{app:kv prompt}
\paragraph{KV retrieval} Here is an example of the context of a KV retrieval task and different types of the appended questions, including direct retrieval, multi-step-question retrieval, multi-matching retrieval, multi-matching retrieval but only need to retrieve the last one, and logic-based tasks:

\begin{tcolorbox}[colback=white,title=Context,colbacktitle=gray,breakable]
\small{{\sffamily
Json data with 3000 key-value pairs:

\{``1532968704'': 78, ``5921306748'': 84, ``3742815096'': 47, ......, ``3276918540'': 76\}
}}
\end{tcolorbox}

\begin{tcolorbox}[colback=white,title=Question,colbacktitle=blue!50!white,breakable]
\small{{\sffamily
\textbf{Direct Retrieval: }In the above json data, please find the value of the key '6978024153'. Give your final answer (the value) in format of ``value: \{answer\}"
\tcbline
\textbf{Reasoning and aggregation (multi-step-question):}\\
Question: In the above json data, please find the value (you need to search it in the Json dictionary) of the Key. The Key is the string S.\\ 
S is the sequential concatenation of A and B. \\
A is the sequential concatenation of the corresponding values (you need to search it in the Json dictionary) of the keys ``8517603942", ``2681307945", ``3759160248", ``6201843957", ``7138095462''. When concatenating, each value is seen as a character.\\
B is a string ``85467''.\\ 
Let's think step by step, and give your final answer (the key and the value) in format of ``key:\{answer\} value:\{answer\}"
\tcbline

\textbf{Multi-matching retrieval:}\\
Question: In the above json data, please find all the keys with the value 0. Give your answer (the keys separated by comma) in format of ``keys: \{answer\}"
\tcbline

\textbf{Multi-matching retrieval (only need to retrieve the last one):}\\
Question: In the above json data, please find all the keys with the value 0. You only need to find one more key to complete the answer.
Answer: The 3 keys whose value is 0 are: \\
key 1: ``1864209357"\\
key 2: ``3145286970"\\
key 3: \{answer\}
\tcbline

\textbf{Logic based retrieval:}\\
Question: In the above json data, please find the Key (only one) whose Value (an integer) is greater than 223 and smaller than 278. Give your answer (the key) in format of ``key: \{answer\}"}}
\end{tcolorbox}

\paragraph{Student resume retrieval} Here is an example of the student resume retrieval task. All the students information are converted to declarative sentences of a fixed format in the context.

\begin{tcolorbox}[colback=white,title=Context,colbacktitle=gray,breakable]
\small{{\sffamily
Here are 100 students' resumes:\\ 

The student named Hallie Turner is 21 years old, graduated from New York University with a GPA of 4.96. He/She is interested in basketball and his/her self-introduction is: Creative writer exploring the impact of social media on culture.\\

The student named Sonali Jain is 22 years old, graduated from Mithibai College with a GPA of 2.52. He/She is interested in Writing and his/her self-introduction is: An artist and writer inspired by travels across cultures.\\

The student named ......}}
\end{tcolorbox}

\begin{tcolorbox}[colback=white,title=Question,colbacktitle=blue!50!white,breakable]
\small{{\sffamily
\textbf{Simple retrieval:}\\
Question: What is the age of the student named Hallie Turner?\tcbline

\textbf{Multi-matching retrieval:}\\
Question: Please find all the students who graduated from Tokyo University of Agriculture and Technology. Please give your final answer (the students' names separated by commas) in the format of ``names: \{answer\}"\tcbline

\textbf{Logic based retrieval (arithmetic):}\\
Question: Which student has a GPA between 2.25 and 2.58? Please give your final answer (the student's name) in the format of ``name: \{answer\}"}

\textbf{Logic based retrieval (classification):}\\
Question: Which student's interest belongs to Sports? Please give your final answer (the student's name) in the format of ``name: \{answer\}"}
\end{tcolorbox}

\section{Complete Evaluation Results}
\label{full result}

In this section, we present our complete evaluation results of various models on 4 types of retrieval tasks: direct retrieval, multi-step-question retrieval, multi-matching retrieval and logic-based retrieval. The data and prompt we used are shown in Appendix \ref{sec:data}.

\begin{table*}[htb]
    \centering
\begin{tabular}{ll|cccc}
\toprule
Task & Model & $N$=10~ & $N$=100 & $N$=1000 & $N$=3000 \\
\midrule
\multirow{3}{*}{k-v} & phi-3.5 & 100 & 99 & 92 & 85 \\
                      & llama3.1 70b & 100 & 100 & 99 & 100 \\
                     & deepseek & 100 & 100 & 100 & 100 \\
                      & gpt-4o & 100 & 100 & 100 & 100 \\
\midrule
\multirow{3}{*}{v-k} & phi-3.5 & 98 & 67 & 7 & 0 \\
                      & llama3.1 70b & 100 & 100 & 99 & 100 \\
                      & deepseek & 100 & 99 & 10 & 0 \\
                      & gpt-4o & 100 & 100 & 100 & 100 \\
\bottomrule
\end{tabular}

    \caption{model score on simple KV retrieval tasks with 2 problem types: k-v means given key, search value and v-k means given value, search key. The context lengths with respect to KV number 10, 100, 1000, 3000 are 0.1k,1k,10k,30k.}
  	\label{tab:length increase kv}
\end{table*}

\begin{table*}[htb]
    \centering
    \begin{tabular}{lcccc}
\toprule
Model & Total KVs ($N$) & 1 step & 3 steps & 5 steps \\
\midrule
\multirow{3}{*}{phi-3.5} & 10  & 85, 68 & 58, 49 & 15, 17  \\
 & 100 & 60, 48 & 57, 30 & 3, 13  \\
 & 1000 & 62, 10 & 13, 0 & 0, 7  \\
\midrule
\multirow{3}{*}{llama3.1 70b} & 10 & 99, 95 & 97, 95 & 78, 83  \\
 & 100 & 96, 96 &97, 92 & 71, 78\\
& 1000 & 96, 88 &91, 55 & 50, 50\\
\midrule
\multirow{3}{*}{deepseek} & 10 & 100, 100 & 100, 100 & 100, 100  \\
 & 100 & 100, 93 & 100, 100 & 100, 100\\
& 1000 & 100, 40 & 97,40 & 93,60   \\
\midrule
\multirow{3}{*}{gpt-4o} & 10 & 100, 100  & 100, 100  & 100, 100  \\
 & 100 & 100, 100  & 100, 100  & 100, 100 \\
 & 1000 & 100, 100  & 100, 100  & 94, 97 \\
\bottomrule
    \end{tabular}
    \caption{Model performance on tasks requiring gathering $n$ values to form a new key and then retrieve its value. We record 2 scores, the former is the accuracy of forming the Key and the latter is the accuracy of getting the Value.}
  	\label{tab:multi-step score}
\end{table*}

% table: multi KV
\begin{table*}[htb]
\centering
\begin{tabular}{lc|cccc}
\toprule
Model & Total KVs & 1 match & 5 matches &10 matches&20 matches\\
\midrule
\multirow{3}{*}{phi-3.5} & 10 & 91 (7/0/2) & 36 (5/30/29) & /& / \\ 
 & 100 & 42 (35/2/21) & 0 (0/10/90) & 0 (0/0/100)& 0 (0/0/100) \\
 & 1000 & 1 (7/13/79) & 1 (7/0/87) & 0 (0/7/93)& 0 (0/0/100)\\
\midrule
\multirow{3}{*}{llama3.1 70b} & 10 & 100 (0/0/0) & 99 (1/0/0) &  /& / \\
 & 100 & 99 (0/0/0) & 59 (17/14/8) & 45 (14/17/22)& 24 (12/34/30)\\
 & 1000 & 62 (0/0/0) & 3 (2/47/46) & 0 (0/15/85)& 0 (0/12/88)\\
\midrule
\multirow{3}{*}{deepseek} & 10 & 100 (0/0/0) & 100 (0/0/0) &  /& / \\
& 100 & 95 (3/0/2) & 40 (9/36/15) & 13 (9/42/36)& 3 (3/43/51) \\
& 1000 & 0 (0/100/0) & 0 (0/100/0) & 0 (0/100/0)& 0 (0/100/0)\\
\midrule
\multirow{3}{*}{gemini} & 10 & 100 (0/0/0) & 100 (0/0/0) & /& / \\
& 100 & 100 (0/0/0) & 82 (14/1/3) & 48 (32/12/8) & 35 (14/36/15)\\
& 1000 & 94 (2/4/0) & 66 (3/26/5) & 18 (9/47/26) & 2 (4/45/49)\\
\midrule
\multirow{3}{*}{gpt-4o} & 10 & 100 (0/0/0) & 100 (0/0/0) & /& / \\
& 100 & 100 (0/0/0) & 98 (0/2/0) & 85 (5/5/5) & 50 (5/40/5)\\
& 1000 & 60 (5/35/0) & 45 (5/45/5) & 5 (0/50/45)& 0 (0/17/83)\\
\bottomrule
\end{tabular}

\caption{Accuracy on KV retrieval, requiring retrieve all the keys with the given value, when the number of total KVs and matching KVs are increasing.}
\label{tab: score multi matching kv detail}
\end{table*}

% table: multi student
\begin{table*}[htb]
\centering
\begin{tabular}{lc|cccc}
\toprule
Model & Total Students & 1 match & 5 matches & 10 matches \\
\midrule
\multirow{2}{*}{phi-3.5} & 10 & 98 (1/0/1) & 27 (13/47/13) & 98 (0/2/0) \\
 & 100 & 33 (52/0/15) & 0 (1/4/95) & 0 (0/1/99) \\
\midrule
\multirow{2}{*}{llama3.1 70b} & 10 & 100 (0/0/0) & 99 (0/1/0) & 100 (0/0/0) \\
& 100 & 98 (1/0/1) & 21 (8/48/23) & 0 (3/45/52) \\
\midrule
\multirow{2}{*}{deepseek} & 10 & 100 (0/0/0) & 91 (1/8/0) & 100 (0/0/0) \\
 & 100 & 100 (0/0/0) & 23 (1/54/22) & 0 (1/36/63) \\
\midrule
\multirow{2}{*}{gemini} & 10 & 100 (0/0/0) & 100 (0/0/0) & 100 (0/0/0) \\
& 100 & 96 (4/0/0) & 36 (8/47/9) & 8 (5/37/50) \\
\midrule
\multirow{2}{*}{gpt-4o} & 10 & 100 (0/0/0) & 100 (0/0/0) & 100 (0/0/0) \\
 & 100 & 100 (0/0/0) & 65 (0/20/15) & 0 (0/75/25) \\
\bottomrule
\end{tabular}
\caption{Accuracy on student resume retrieval, requiring retrieve all the students graduating from the given university, when the number of total students and matching students vary.}
\label{tab: score multi matching resume detail}
\end{table*}

\begin{table*}[htb]
\centering
\begin{tabular}{l|cccc|ccc}
\toprule
\multirow{2}{*}{Model} & \multicolumn{4}{c|}{KV Retrieval} & \multicolumn{3}{c}{Student Resume Retrieval} \\
\cmidrule{2-8}
 & $N$=4~~ & $N$=10~ & $N$=100& $N$=1000 & $N$=4~~  & $N$=10~ & $N$=100 \\
\midrule
phi-3.5 & 69 & 9 & 0 & 0 & 75 & 53 & 9 \\
llama-3.1-70b & 72 & 41 & 6 & 1 & 81 & 63 & 13 \\
deepseek & 84 & 67 & 12 & 0 & 94 & 81 & 21 \\
gemini & 78 & 33 & 6 & 0 & 92 & 80 & 12 \\
gpt-4o & 97 & 87 & 38 & 5 & 100 & 92 & 30 \\
\bottomrule
\end{tabular}
\caption{The accuracy (\%) on logic-based KV retrieval and Student Resume Retrieval.}
\label{tab:combined logic}
\end{table*}

\begin{table}[htb]
\centering
\begin{tabular}{l|cccc}
\toprule
Model &  $N$=4  & $N$=10 & $N$=30 &  $N$=100 \\
\midrule
gemini & 88 & 82 & 55 & 20  \\
gpt-4o & 92 & 88 & 72 & 42  \\
\bottomrule
\end{tabular}
\caption{The accuracy (\%) on classification-based logic-based Student Resume Retrieval.}
\label{tab:logic interest}
\end{table}

\subsection{Direct Retrieval: Usually Simple, But Sensitive to Question Types}
\label{app:direct}

Previous benchmarks have demonstrated the model's strong performance in direct retrieval tasks such as NIAH \citep{gkamradt_llmtest_needleinahaystack_2023} and Key-Value retrieval \cite{zhang_inftybench_2024}, even when the context length exceeds 100k tokens. However, interestingly, our experiments with 2 simple types of direct retrieval tasks with only slight differences reveal that some models may be more sensitive to the question type, while the extremely long context itself is not of primary concern.

For KV retrieval task across different context lengths, we test two types of problems: searching for a value given a key (k-v), and searching for a key given a value (v-k). The results, shown in Table \ref{tab:length increase kv}, indicate that in the ``k-v'' task, increased context length does not significantly affect model performance, because the task is inherently simple for the model. However, for phi-3.5 and deepseek, a slight modification in the problem to a ``v-k'' form, while keeping the context length constant, leads to a sharp decline in performance, which is much greater than that caused by simply increasing the context length for the same problem. This suggests that models are more sensitive to the type of problem than to the length of the context. (We infer that models like phi-3.5 and deepseek may treat ``v-k'' task as a numeric comparing task rather than direct retrieval, thus perform very poor in long context.)

Therefore, we conclude, current long-context models can indeed perfectly pass tests like Key-Value retrieval or NIAH \citep{gkamradt_llmtest_needleinahaystack_2023} in 128k length, but they may be highly susceptible to the specific form of questions, even when the context itself remains unchanged. So we speculate that problems with seemingly similar forms may have fundamentally different natures.

% the context length is not the primary factor determining the difficulty or performance issues of powerful models like GPT-4 on long context tasks. Instead, the nature of the problem itself plays a more significant role. 

\subsection{Multi-step-question Retrieval: Can be Solved By Normal CoT}
\label{app:multi step}

Multi-step-question retrieval tasks, including multi-query, multi-hop, chain-of-retrieval tasks, etc. \citep{li_needlebench_2024,wang_leave_2024}, intuitively appear more challenging due to the need to aggregate dispersed information from the context by multiple steps. Nevertheless, our experiments indicate that they are not necessarily difficult, provided LLMs can automatically decompose them into simpler steps using CoT approach \citep{wei_chain--thought_2023}. We must emphasize that, it is important to distinguish multi-matching from multi-step-question retrieval: the former requires retrieving multiple items with only one query, while the latter also requires retrieving multiple items but can be achieved with multiple different queries.

Previous studies \citep{wang_leave_2024, goldman_is_2024, li_needlebench_2024, li_loogle_2023} typically characterize long-context tasks, which require the aggregation of dispersed information for multi-step reasoning, as difficult by default. While this assessment aligns with intuition, we argue that this classification is not universally scientific. Instead, the complexity of such tasks should be evaluated based on their specific composition, because a complex task may not be inherently difficult if a model can systematically reason and decompose the task into manageable steps.

To illustrate this, we constructed a multi-step-question Key-Value retrieval problem that involves a chain-of-retrieval process. This task requires a model to retrieve $n$ values corresponding to $n$ keys, concatenate them into a string, and then combine this string with another given string to generate a new key, from which the model must retrieve its value as the final answer (see detailed prompts in Appendix \ref{app:kv prompt}). Although this problem seems complex, requiring aggregating at least $n$+1 pieces of information from different parts of the context and performing at least $n$+1 reasoning steps, it can actually be decomposed into simple steps using a chain-of-thought (CoT) approach \citep{wei_chain--thought_2023}.

The results presented in Table \ref{tab:multi-step score} reveal that most models's performance are almost unaffected except the smallest one, phi-3.5, as the numbers of reasoning steps and total items increase. And we find the models automatically adopt CoT in most cases. That means LCLM with good comprehension, reasoning, and organizational abilities can easily solve them.

% Therefore, we conclude that the need for reasoning is not the primary factor contributing to the suboptimal performance of powerful models like GPT-4 on long-context tasks. The difficulty of multi-step problems serves more as a test of a model's abilities in understanding, decomposing, and reasoning about tasks, rather than a capability unique to long-context scenarios.

% large models can performs well as long as using CoT \citep{wei_chain--thought_2023}, even though the context length is longer. Small models performs worse, maybe because its weak reasoning and comprehension capability.

\subsection{Multi-matching Retrieval: Difficult to Provide a Complete Answer}
\label{app:detail multi match}

For multi-matching retrieval, the model generates a list of items as the answer, which can be categorized into four distinct cases:

\begin{enumerate}
    \item \textbf{Fully Correct}: The model's response exactly matches the correct answer, i.e., both sets are equal.
    \item \textbf{Over-selection}: The correct answer is a proper subset of the model's response.
    \item \textbf{Under-selection}: The model's response is a proper subset of the correct answer.
    \item \textbf{Mis-selection}: The model's response and the correct answer do not overlap as subsets of each other.
\end{enumerate}

In Table \ref{tab: score multi matching kv detail} and Table \ref{tab: score multi matching resume detail}, we detail the model's performance, with the first number representing the rate of fully correct responses, while the numbers in parentheses denote over-selection, under-selection, and mis-selection, respectively.

The results reveal that when only 1 matching item is present, larger models, such as Gemini \citep{gemini_team_gemini_2024}, demonstrate superior performance, achieving an accuracy of up to 94\% even in lengthy contexts. However, with the introduction of multiple matching items, such as 5 or 10, the accuracy of all language models rapidly declines to nearly zero, particularly evident in the more realistic scenario of Student Resume Retrieval. This trend suggests that the inherent difficulty of the task is consistently challenging across models of varying sizes.

\subsection{Logic-based Retrieval: Difficult}

The result of arithmetical logic-based retrieval are illustrated in Table \ref{tab:combined logic}, and that of classification-based logic-based retrieval is in \ref{tab:logic interest}. When the context contains only 4 items, LLMs successfully select the correct item in most cases, indicating that these models possess logical judgment capabilities. However, for both datasets, all tested models struggle with these tasks as the context length increases, consistently retrieving an incorrect item whose value lies outside the specified range. It is reasonable to anticipate that these difficulties will be further exacerbated in more complex scenarios that require logical judgment beyond mere numerical comparison and simple classification.

\section{Analytical Experiments}
\label{analytical}

To delve deeper into why LLMs struggle to solve these seemingly simple tasks, we conduct some analytical experiments, including studying LLMs' hidden states and attention weights, to observe LLMs' behavior when handling difficult retrieval tasks. We aim to answer the question: Do these difficult retrieval tasks fundamentally differ from traditional retrieval tasks? 

Specifically, our findings are as follows, which will be detailed in the following sections. Some of them may be trivial.

For logic-based retrieval:
\begin{enumerate}
    \item The internal behavior of LLMs in logic-based retrieval is more akin to arithmetic tasks (i.e. logical tasks) which necessitate reasoning with multiple steps.
    \item Logic-based retrieval can be decompose to 2 components: decide which value is in the range, and then get the corresponding key. However, vector retrieval cannot even solve the first one if within one step.
\end{enumerate}

And for multi-matching retrieval:
\begin{enumerate}
    \item The model's internal behavior in multi-matching retrieval is originally an one-by-one retrieval process.
    \item Even when we decompose a multi-matching retrieval problem into $n$ single-item retrievals, the difficulty of retrieval still continuously increases for items searched later. In other words, it is still not easy to retrieve just one of the matching items in a multi-matching retrieval task.
\end{enumerate}

\subsection{Experiment Settings for Hidden States and Attention Analysis}
\label{probing setting}

In our experiments, analyzing the internal behavior of the model is the most crucial. We employ linear probing to explore the information encapsulated in hidden states and statistically analyze attention weights to infer the model's internal mechanisms. Given our earlier observations of similar performance trends between large and small models, we adopt the lightweight small model, phi-3.5-mini \citep{abdin_phi-3_2024}, which consists of 32 layers and has a hidden size of 3072, for simplicity.

The dataset used is still Key-Value retrieval, but more simplified: the total number of KVs does not exceed 100, the value range is constrained to 0 to 9, and we do not apply the chat template to allow the model to generate answers directly following the question.

In both hidden state and attention analyses, we first identify the anchor token, whose hidden states are used for linear probing and function as the query token in the attention mechanism. In tasks where $n$ is 1, the last token of the question serves as the anchor token. In multi-matching retrieval tasks, $n$ is set to 3, and the 3 gold Keys are appended to the question. The token immediately preceding each of the 3 appended Keys acts as the anchor token, i.e., we conduct experiments on 3 anchor tokens separately. 
The selected anchor tokens in different question types are highlighted in red in the examples below:

\begin{tcolorbox}[colback=white,title=Anchor Tokens,colbacktitle=red!80!white]
\small{{\sffamily
\textbf{Direct retrieval:} \{Context\} In the above JSON data, the Key whose Value is 5 is: \textbf{\textcolor{red}{``}}\\

\textbf{Logic-based retrieval:} \{Context\} In the above JSON data, the Key whose Value is larger than 4 and smaller than 6 is: \textbf{\textcolor{red}{``}}\\

\textbf{Multi-matching retrieval:} \{Context\} In the above JSON data, all the Keys whose Value is 5 are: \textbf{\textcolor{red}{``}}1532968704'', \textbf{\textcolor{red}{``}}5921306748'', \textbf{\textcolor{red}{``}}3742815096''}}
\end{tcolorbox}

In examining hidden states, we employ linear probing to determine whether the model has successfully retrieved the correct Key and stored its information in the hidden states output by each layer. The label, i.e. the probing target, is the first digit of the gold Key to be retrieved, enabling our linear probe to function as a 10-class classifier with a single linear layer. We use 1,600 samples for training and 400 for testing the classifier, and for each layer, we train and test the classifier independently. We train it for 8 epochs with the learning rate of $10^{-5}$.

To elucidate the model's attention dynamics, we compute ``relative attention'' which the anchor token pays to the gold Key and Value in each layer. ``Relative attention'' is the ratio of the average attention weight directed towards the gold Key (or Value) to that towards all other candidate Keys (or Values), reflecting the model's capacity to focus on the target one and exclude distractors. 
%The absolute attention weights are shown in Appendix \ref{app:3 key 17}.

%探究逻辑问题
\subsection{Analyze Logic-Based Retrieval}

\subsubsection{Internal Model Behavior: More Like Multi-Step Arithmetic}
\begin{figure*}[htb]
    \centering
    \includegraphics[width=0.8\linewidth]{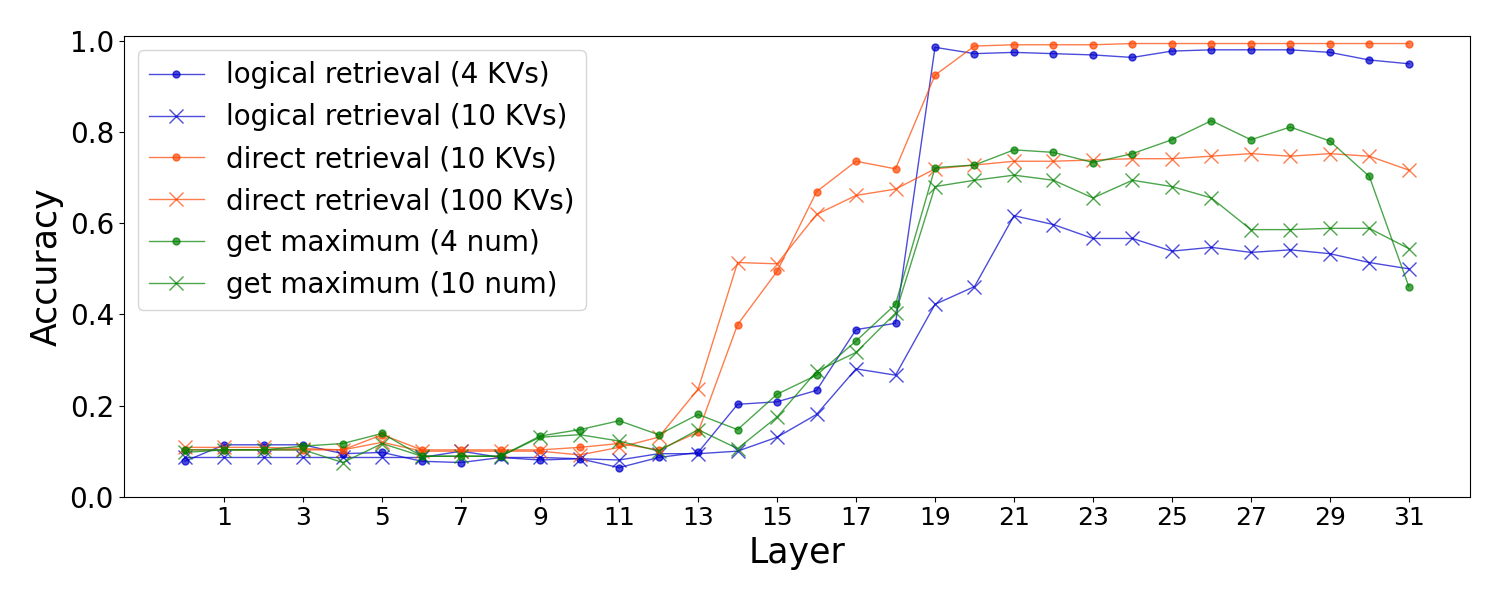}
    \caption{Linear probing accuracy in each layer of 3 tasks: direct KV retrieval, logical KV retrieval, and getting the maximum value among $N$ numbers.}
    \label{fig:probing logic kv}
\end{figure*}

First, we analyze the process of a logic-based KV retrieval task, comparing it with two other tasks: (1) direct KV retrieval, where the model retrieves the Key corresponding to a given Value, and (2) a basic arithmetic task that involves identifying the largest number among $N$ integers ranging from 0 to 100 (see the problem settings in Appendix \ref{app:numeric comparison}). In task (2), we use the ground-truth's unit's digit as the label for probing, and the last token of the prompt as the anchor token.

\begin{figure}[htb]
    \centering
    \includegraphics[width=0.98\linewidth]{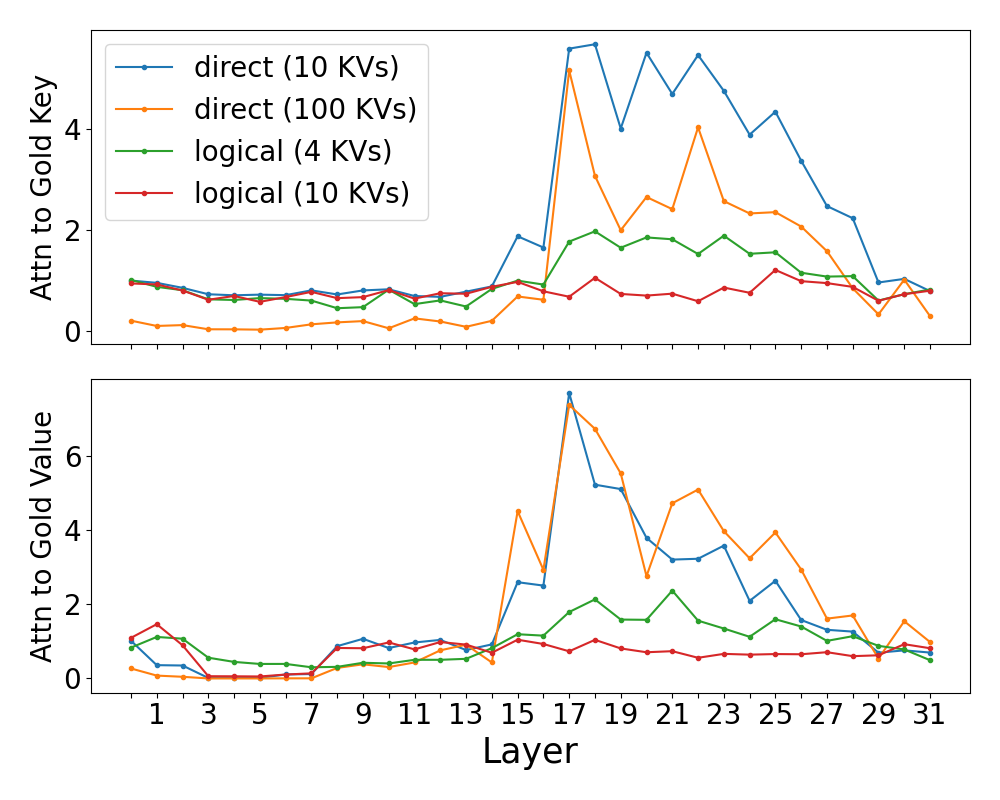}
    \caption{Relative attention to the gold Key and Value of each layer of different KV retrieval tasks.}
    \label{fig:attn logic kv}
\end{figure}

From the linear probing accuracy (Figure \ref{fig:probing logic kv}) and the attention dynamics (Figure \ref{fig:attn logic kv}), we find that model behavior in logic-based retrieval is fundamentally different from standard retrieval and more resembles the numeric comparison task:

1. For direct retrieval, probing accuracy increases in discrete jumps, notably between layers 14 and 20, rather than progressively across all layers. The relative attention toward the gold Key and Value also peaks between layers 15 and 23, suggesting that this attention correlates with the retrieval behavior. This indicates that retrieval activity is concentrated within particular layers (e.g. layers 14 and 20).

2. In logic-based retrieval, the attention curves show a distinct trend and remain consistently lower than direct retrieval. What's more, the point at which the accuracy of logic-based retrieval begins to improve, layer 19, is noticeably later than that of direct retrieval, layer 14; instead, it more closely mirrors the numeric comparing task, which also begins to increase at layer 19. This indicates that logic-based retrieval problems differ fundamentally from normal retrieval tasks, bearing more resemblance to arithmetic problems.

A prior research \citep{feng_towards_2023} has theoretically demonstrated that a constant-size transformer model cannot solve arithmetic problems with many steps in a single step unless using CoT \citep{wei_chain--thought_2023}. Therefore, a logic-based retrieval, as it resembles arithmetic, will also be unsolvable in one step as long as $N$ is large enough.

%problem whose context consists of $N$ items will require $N$ logical judgments, thus always 
\subsubsection{Can Vector Retrieval Solve Logic-based Retrieval?}
Second, we intuitively decompose a logic-based retrieval problem into 2 components: decide which value is in the range, and get the corresponding key. Since our focus is on retrieval scenarios, we directly employ models designed for RAG \citep{lewis2020retrieval} to test the feasibility of performing the first step through vector retrieval. (The second step is the same as normal retrieval.)

Vector retrieval is one of the most widely retrieval techniques, which encodes both the query and candidates into embedding vectors, then calculates the similarity between the query and each candidate to retrieve the top similar options. This process is similar to the attention mechanism within LLMs \citep{xu_retrieval_2023}. Since similarity computation can be conducted in parallel for each candidate, it can be considered as a single-step process from an external perspective.

We test 2 sentence embedding models commonly used for RAG \citep{lewis2020retrieval}, e5-large-multilingual \citep{wang_multilingual_2024} and bge-m3\citep{zhang_retrieve_2023}, on a numerical value comparing task to see if it can retrieve the correct number from 20 integers. The candidate keys are integers within the range 0 to 30, 100, 1,000, or 10,000, and the queries have 2 types, equality relations and greater\&less-than comparison (examples are shown in Appendix \ref{app:embed}). If the embedding vector of the correct number has the highest similarity to that of the query, it is considered correct. As shown in Table \ref{tab:embed model}, they only function properly when retrieval is based on equality relations; however, their performance significantly declines with more advanced logical operations, such as greater-than or less-than comparisons. 

This suggests that one-step vector retrieval techniques are inadequate for logic-based retrieval. In other words, a transformer model cannot achieve logic-based retrieval through the attention mechanism within a single layer (or a few layers); instead, it requires a more advanced reasoning process.

\begin{table*}[h]
\centering
\begin{tabular}{lc|ccccc}
\toprule
Model & Criteria Type & within 30 & within 100 & within 1k & within 10k \\
\midrule
\multirow{2}{*}{e5-large} & Greater/Less Than & 36 & 31 & 21 & 16 \\
 & Equal & 95 & 98 & 99 & 99 \\
\midrule
\multirow{2}{*}{bge-m3} & Greater/Less Than & 37 & 29 & 20 & 23 \\
 & Equal & 95 & 98 & 99 & 100 \\
\bottomrule
\end{tabular}
\caption{Accuracy of sentence embedding models on numerical comparison retrieval, with 2 different types of retrieval criteria, as the range of random selected numbers increases. The number of test samples are 100.}
\label{tab:embed model}
\end{table*}

%证明多匹配是多步骤
\subsection{Analyze Multi-Matching Retrieval}

\subsubsection{Internal Model Behavior: Retrieve One-by-one But Increasingly Hard for Retrieving Later Items}

\begin{figure*}[htb]
    \centering
    \subfloat[Linear probing\label{fig:probing 3 matching}]{
    \includegraphics[width=0.55\linewidth]{figures/probing_acc_3match.png}
    }
    \subfloat[Relative attention\label{fig: pred 3 key attn}]{
    \includegraphics[width=0.45\linewidth]{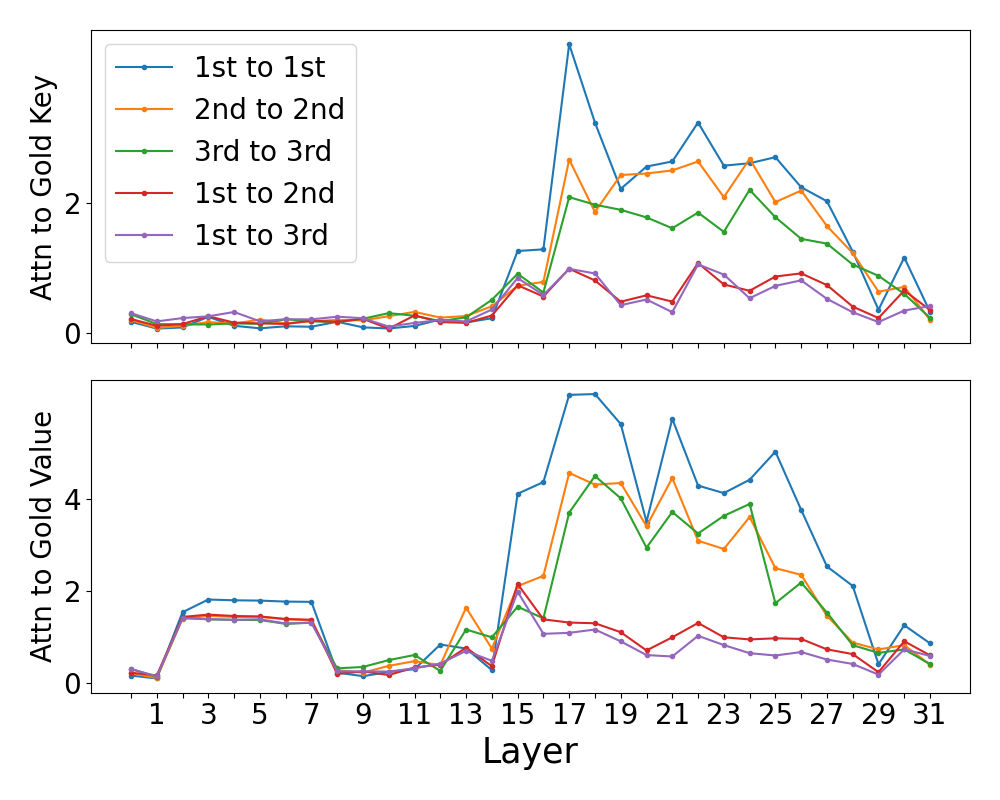}
    
}
    \caption{(a) The accuracy of probing from hidden states of each layer when predicting the first, second, and third Key. For example, ``2nd key at 1st'' means we probe from the hidden states of the 1st anchor token but use the 2nd anchor token (the 2nd gold Key's first digit) as the label for training and testing. (b) Attention to each Key or Value when predicting the first, second, and third Key respectively. For example, ``1st to 2nd'' means we calculate the attention to the 2nd gold Key (or Value) with the 1st anchor token as the query token.}
 \label{fig: pred 3 key attn and probing}
 \end{figure*}

For the multi-matching situation, first, we examine the model's behavior when tasked with predicting each matching item. The relative attention curve (Figure \ref{fig: pred 3 key attn}) and the linear probing accuracy (Figure \ref{fig:probing 3 matching}) demonstrate that:
%We find that predicting later items is increasingly harder for LLMs, despite there in fact being no priority relationship among the items.

1. In multi-matching retrieval, the model retrieves the matching items one by one rather than simultaneously, as no information about the 2nd or 3rd key can be detected at the end of the question (i.e. the first anchor token), which proves the model does not retrieve multiple items all at once. Correspondingly, the attention to the 2nd or 3rd key at the end of the question is very low, and the model mainly focus on only the 1st key.

2. The model can accurately retrieve the first Key, akin to the behavior in direct retrieval. However, for the subsequent Keys, both attention and probing accuracy are much lower, with the 3rd Key being harder to retrieve than the 2nd. This proves predicting later items is increasingly harder for LLMs, despite there in fact being no priority relationship among the items.

\subsubsection{Is Retrieving Just One Item Simple?}

\begin{table*}[h]
\centering
\begin{tabular}{lc|cccc}
\toprule
Model & Total KVs & 3 matches & 10 matches & 30 matches & 100 matches \\
\midrule
\multirow{2}{*}{phi-3.5} & 100 & 22 & 2 & 0 & / \\
 & 1000 & 0 & 0 & 0 & 0 \\
\midrule 
\multirow{2}{*}{llama3.1 70b} & 100 & 91 & 66 & 54 & / \\
 & 1000 & 61 & 43 & 16 & 1 \\
 \midrule 
 \multirow{2}{*}{deepseek} & 100 & 93 & 75 & 37 & / \\
 & 1000 & 15 & 5 & 6 & 0 \\
 \midrule 
\multirow{2}{*}{gpt-4o} & 100 & 100 & 92 & 42 & / \\
 & 1000 & 95 & 42 & 25 & 1 \\
 \midrule 
\multirow{2}{*}{gemini} & 100 & 100 & 99 & 67 & / \\
 & 1000 & 99 & 90 & 49 & 8 \\
\bottomrule
\end{tabular}
\caption{Accuracy on KV retrieval, requiring retrieving only the last Key (of $n$ Keys) with the given Value, and other Keys with the given Value are already given.}
\label{tab:only one}
\end{table*}

Second, to further prove that retrieving a later item is indeed more challenging, we design a simplified version of the multi-matching Key-Value (KV) retrieval problem: the model is provided with all but one of the $n$ matching Keys and is required to predict the last remaining Key (see detailed prompts in Appendix \ref{app:kv prompt}). The Key to be retrieved is selected randomly, meaning it may not necessarily be the last one that appears in the sequence of the input context.

The results in Table \ref{tab:only one} demonstrate that even when the task is ostensibly reduced to predicting a single item, corresponding to just one step in the one-by-one retrieval process, the difficulty still increases to very high as the number of matching items grows. The results indicate that retrieving even a single item is not simple, if the item is a later (the order in which items appear in the output sequence, rather than the input sequence) retrieved one.

We speculate the increasing difficulty may stem from the increasing complexity of retrieval criteria for later items. Retrieving subsequent Keys requires more extensive and stringent criteria, including conditions to exclude previously retrieved items, thereby complicating the retrieval process. In other words, the model need to pay more efforts to exclude previously retrieved items to avoid getting too many items at the same time. Therefore, multi-matching retrieval also has the multi-step nature: it not only needs $n$ steps to generate $n$ items for $n$-matching retrieval, and but also may need additional $k$-1 steps to exclude previous ones for the $k$-th item. Consequently, the number of steps required may be proportional to $n^2$.

\subsection{Additional Information of Problem Settings}
%数值比较任务
\subsubsection{Arithmetic Problem}
\label{app:numeric comparison}
Finding the biggest number among $N$ numbers typically needs $N$ steps, which belongs to the simplest multi-step logical problems. Our prompt is as the following, where the model will predict the correct answer directly after the prompt.

\begin{tcolorbox}[colback=white]
A list of integers: 15, 24, 31, 44. In the list, the biggest integer is  
\end{tcolorbox}

%rag模型
\subsubsection{Sentence Embedding Models on Math problems}
\label{app:embed}
We test 2 commonly used sentence embedding model, e5-large-multilingual \citep{wang_multilingual_2024} and bge-m3\citep{zhang_retrieve_2023}, on a numerical value comparing task to see if it can retrieve the correct number from 20 integers. The candidate keys are integers within the range 0 to 30, 100, 1,000, or 10,000, and the queries have 2 types: (1) equality judgement (2) greater\&less than judgement. As the example:
\begin{tcolorbox}[colback=white,title=Query and Candidates,colbacktitle=ProcessBlue,breakable]
\small{{\sffamily
\textcolor{JungleGreen}{\textbf{query (equal)}}: The integer equal to 310.\tcbline
\textcolor{red}{\textbf{query (greater than and less than)}}: The integer smaller than 223 and larger than 356.\tcbline
\textbf{20 candidates (range 0-1000)}: \textcolor{red}{310}, 734, 296, 501, 893, 178, 645, 923, 57, 782, 464, 852, 213, 689, 371, 970, 510, 116, 455, 682}}
\end{tcolorbox}

\section{Using Specific CoT-like Prompts}
\subsection{Examine Each Item One-by-one to Retrieve}
\label{app:one by one}
Taking a logic-based KV retrieval task for example, when given a normal question without CoT-like prompt, the model (gpt-4o) tends to make mistakes:

\begin{tcolorbox}[colback=white,title=Context,colbacktitle=gray,breakable]
\small{{\sffamily
Json data with 100 key-value pairs:\\

\{``0214587963'': 933, ``9578042316'': 354, ``8593746120'': 143, ``9473061852'': 353, ...... ``0587214936'': 448, ``6742193850'': 379, \textbf{``2405163897'': 572}, ``4871369052'': 768,  ......

 ``6028917354'': 448, ``3249758610'': 938\}}}
\end{tcolorbox}

\begin{tcolorbox}[colback=white,title=Standard Q\&A,colbacktitle=blue!50!white,breakable]
\small{{\sffamily
\textbf{Question: }
In the above json data, please find the Key (only one) whose Value (an integer) is greater than 527 and smaller than 620. Give your answer (the key) in format of ``key: \{answer\}"\tcbline

\textbf{Wrong Answer: }
The key whose value is greater than 527 and smaller than 620 is ``1748263950'' with a value of 525.}}
\end{tcolorbox}

If we use a special CoT-like prompt to let it to examine each item one by one, it generates about 2000 tokens to judge every KV pair, which takes a long time but finally gives the right answer (for multi-matching problems, the process is similar):

\begin{tcolorbox}[colback=white,title=Check Each item One-by-one Q\&A,colbacktitle=orange!80!white,breakable]
\small{{\sffamily
\textbf{Question: }
In the above json data, please find the Key (only one) whose Value (an integer) is greater than 527 and smaller than 620. \\ 

\textbf{Please check each item one by one and return your judgment (yes/no) on whether it meets the requirements.} Give your final answer (the key) in format of ``key: \{answer\}"

\tcbline

\textbf{Correct Answer:}
To find the key whose value is greater than 527 and smaller than 620, let's examine each key-value pair in the provided JSON data:

1. ``0214587963'': 933 (no)\\
2. ``9578042316'': 354 (no)\\
......\\
50. ``6742193850'': 379 (no)\\
51. ``2405163897'': 572 (yes)\\
52. ``4871369052'': 768 (no)\\
......\\
100. ``3249758610'': 938 (no)\\

The key that meets the requirement is ``2405163897'' with a value of 572.

Final answer: key: 2405163897}}
\end{tcolorbox}

\subsection{Examine Each Item One-by-one And Maintain A List}

Taking student resume analysis for example, when asking the model to update a list while examining each item, the prompt design and the answer of gpt-4o are as follows:

\begin{tcolorbox}[colback=white,title=Check Each item One-by-one and Add to a List Q\&A,colbacktitle=orange!100!white,breakable]
\small{{\sffamily
\textbf{Question: }
Please find all the students who graduated from Jilin University. Please give your final answer (the students' names separated by commas) in the format of "names: {answer}" \\

\textbf{To solve this question, please do the following steps directly by yourself (instead of writing code):\\
1. First, initialize an empty list to store the matching items.\\
2. Sequentially examine every item one by one. if it meet the requirement, output "true" and add it to the list, and print the current list; otherwise, output "false" and just continue to the next one.\\
3. Finally, according to the list, summarize your final answer to the question.\\}

\tcbline

\textbf{Correct Answer:}
1. Initialize an empty list: \`matching\_students = []\`.\\

2. Sequentially examine each student:\\

- Kieran Adams: Graduated from Jilin University. True. Add to list. Current list: \`['Kieran Adams']\`.\\
- Xie Yan: Graduated from Xiamen University. False. Continue.\\
- Leo Hall: Graduated from University of Pennsylvania. False. Continue.\\
- Rika Sakamoto: Graduated from Jilin University. True. Add to list. Current list: \`['Kieran Adams', 'Rika Sakamoto']\`.\\
......\\

3. Final answer: names: Kieran Adams, Rika Sakamoto, Emiko Fujiwara, Shiori Yoshida, Mia Garcia, Deng Xin, Ava Martinez, Aditya Bhat, Serena Morgan, Noah Lewis}}
\end{tcolorbox}

\subsection{Use Program of Thoughts to Retrieve}
\label{app:PoT}
\begin{tcolorbox}[colback=white,title=Program of Thoughts Q\&A,colbacktitle=JungleGreen!60!white,breakable]
\small{{\sffamily
\textbf{Question: }
In the above json data, please find the Key (only one) whose Value (an integer) is greater than 527 and smaller than 620.  You can write a python program to solve this, and regard the json data as a file ``KV.json''.\\

\textbf{Answer:}
To find the key whose value is greater than 527 and smaller than 620, you can use the following Python program. This program reads the JSON data from a file named ``KV.json'' and searches for the key that meets the criteria.

\begin{lstlisting}
import json

# Load the JSON data from the file
with open("KV.json", ``r") as file:
    data = json.load(file)

# Initialize a variable to store the desired key
desired_key \= None

# Iterate over the key-value pairs in the JSON data
for key, value in data.items():
    # Check if the value is greater than 527 and smaller than 620
    if 527 < value < 620:
        desired_key = key
        break  # Since we only need one key, we can break once we find it

# Print the desired key
print("The key with value greater than 527 and smaller than 620 is:", desired_key)

\end{lstlisting}
This program will output the key that satisfies the condition. When you run it with the provided JSON data, it will find and print the key `"2405163897"`, which has the value `572`.}}
\end{tcolorbox}

We can also use program-of-thoughts prompt \citep{chen_program_2023} to let the model write programs to solve this retrieval task. Writing python program to solve this task is very easy for gpt-4o, and then by executing this program in an external interpreter, we can get the right result. Here is an example of logic-based KV retrieval (it can similarly easily solve multi-matching problems), whose context is the same as the previous section.

\section{Analyse the Components of Challenging Tasks from Various Long-context Benchmarks}
\label{app:previous benchmark}
% % l-eval longbench loogle 
% % haystack reason, haystack summary, loogle, loong, ruler
% longbench 6 tasks, not hard : multi-doc. single-doc. synthetic, summary,code,few-shot
% bamboo: real-world: question answering, hallucination detection, text sorting, language modeling, and code completion

% loogle difficult: Multiple information retrieval , computation, timeline reorder
% loong: clustering, comparison, chain of reasoning
% ruler: Multi-values NIAH, Multi-queries, Multi-hop Tracing, Aggregation
% BABILong: 20 reasonging task
% longICLbench: Extreme-label In-context Learning
% needle bench: Multi-Needle Retrieval Task (M-RT),Multi-Needle Reasoning Task(M-RS),Ancestral Trace(祖先关系推理)
% summary haystack: summary

\subsection{Distinguishing what is multi-matching or logic-based retrieval}
\label{app:distinguish}

We regard multi-matching and logic-based retrieval as important individual components with distinct natures. It may be difficult for first readers to distinguish between multi-matching and multi-query, or logic-based retrieval and problems requiring logic. Here we use some simple example to illustrate this. The primary characteristic of multi-matching or logic-based retrieval is the inability to be further divided into multiple manageable steps using CoT, unless examining each item in the context one by one.

Multi-matching means multiple items meet one retrieval criteria, which can almost no longer be subdivided into multiple independent conditions. However, as for multi-query, although the retrieval criteria is also in one sentence, it can be easily subdivided. 

\begin{tcolorbox}[colback=white,title=Context,colbacktitle=gray]
\small{{\sffamily
$\star$ Jack is 30 years old.\\
$\star$ Mike is 20 years old.\\
$\star$ Lee is 50 years old.\\
$\star$ William is 20 years old.\\
$\star$ James is 35 years old.}}
\end{tcolorbox}

Here is the example, where the multi-query retrieval question can be subdivided into 3 questions to retrieve the 3 people individually, while the multi-matching retrieval problem cannot.

\begin{tcolorbox}[colback=white,title=Question Differentiation (1),colbacktitle=blue!40!white]
\small{{\sffamily
\textcolor{JungleGreen}{\textbf{Multi-query retrieval (easy):}} How old is Jack, Lee and James respectively? \\ \\
\textcolor{JungleGreen}{\textbf{CoT}}: (1) retrieve Jack's age  (2) retrieve Lee's age  (3) retrieve James's age  (4) summary
\tcbline
\textcolor{red}{\textbf{Multi-matching retrieval (hard):}} Who are 20 years old ?}}
\end{tcolorbox}

Here is the example, where a problem requiring logical reasoning can be easily devided into simple 4 steps, while the logic-based retrieval problem cannot. 

In our study, ``logic" typically specifically refers to logical judgments which cannot simply be reflected by similarity, such as determining that 50 is greater than 45. Although ``equality" is also a logical relationship in mathematics, the equivalence of two identical numbers can always be established through low-level abstract similarity alone. Therefore, LLMs or embedding models do not require advanced ``logic" to judge equality relationships. Therefore, retrieval based on equality is not considered a logic-based retrieval problem in our study.

\begin{tcolorbox}[colback=white,title=Question Differentiation (2),colbacktitle=blue!40!white]
\small{{\sffamily
\textcolor{JungleGreen}{\textbf{Requiring logical reasoning (easy):}} Whose age is equal to the age of Lee minus the age of Jack? 
\\ \\
\textcolor{JungleGreen}{\textbf{CoT}}: (1) retrieve Lee's age (2) retrieve Jack's age  (3) do a simple subtraction and get the target age  (4) retrieve by the target age
\tcbline
\textcolor{red}{\textbf{Logic-based retrieval (hard):}} Who is older than 45?}}
\end{tcolorbox}

\subsection{Analyse Advanced Long-context Tasks}
\label{app:task belong}
\begin{table*}[htbp]
    \centering
    \begin{tabular}{ll|c|ccc}
        \toprule
        \textbf{Benchmark} & \textbf{Task Name}&hard&multi-step-Q&multi-match&logic-based\\
        \midrule
        \multirow{4}{*}{Loogle} 
            & Multiple info retrieval & \faCheckG & \faTimesR & \faCheckG & \faTimesR\\
            & Computation & \faCheckG & \faCheckG & \faTimesR & \faQuestionCircleO\\
            & Timeline reorder & \faCheckG & \faCheckG & \faTimesR & \faCheckG\\
            & Comprehension\&reasoning & \faCheckG & \faCheckG & \faQuestionCircleO & \faQuestionCircleO\\
        \midrule
        \multirow{5}{*}{Ruler} 
            & Multi-keys NIAH & \faTimesR & \faTimesR & \faTimesR & \faTimesR\\
            & Multi-values NIAH & \faTimesR & \faTimesR & \faCheckG & \faTimesR\\
            & Multi-queries NIAH & \faTimesR & \faCheckG & \faTimesR & \faTimesR\\
            & Multi-hop Tracing & \faTimesR & \faCheckG & \faTimesR & \faTimesR\\
            & Aggregation & \faTimesR & \faCheckG & \faCheckG & \faTimesR\\
        \midrule
        \multirow{3}{*}{NeedleBench} 
            & Multi-Needle Retrieval & \faTimesR & \faCheckG & \faTimesR & \faTimesR\\
            & Multi-Needle Reasoning & \faTimesR & \faCheckG & \faTimesR & \faTimesR\\
            & Ancestral Trace & \faTimesR & \faCheckG & \faTimesR & \faTimesR \\
        \midrule
        \multirow{4}{*}{Loong} 
            & Spotlight Locating & \faTimesR & \faTimesR & \faTimesR & \faTimesR\\
            & Comparison & \faCheckG & \faTimesR & \faQuestionCircleO & \faCheckG\\
            & Clustering & \faCheckG & \faTimesR & \faQuestionCircleO & \faCheckG\\
            & Chain of Reasoning & \faTimesR & \faCheckG & \faQuestionCircleO & \faQuestionCircleO\\
        \bottomrule
    \end{tabular}
    \caption{Some advanced long context tasks from different benchmarks. We mark whether the task is very hard, involves multi-step-question or multi-matching retrieval or logic-based retrieval. \faQuestionCircleO \ means this is uncertain, depending on more specific scenarios.} 
    \label{tab:tasks}
\end{table*}

In Table \ref{tab:tasks}, we list some advanced long-context tasks from previous benchmarks, and identify if the task is very hard, involves multi-step-question or multi-matching retrieval or logic-based retrieval.

If the evaluation results from this benchmark indicate that no existing model is able to score above 60 on this task, we will categorize the task as very difficult; otherwise, it will be marked as not very difficult. How we decide whether it is multi-step-question, multi-matching or logic-based is shown in Appendix \ref{app:detail each task}.

Note that multi-matching will not be very difficult if the amount of matching items is not so large, for example, a 3-matching retrieval may be easy for Gemini, but 30-matching must be really hard.

\subsection{Details of these tasks}
\label{app:detail each task}
Here we show the process that we analyse previous challenging tasks from different benchmarks to identify which component these tasks involve. We omit simple tasks like single-needle NIAH or normal multi-document QA.

\subsubsection{Loogle}
Loogle \citep{li_loogle_2023} is a long-context benchmark which first distinguish short-dependency tasks and long-dependency tasks. Long-dependency means the task needs a large portion of the context rather than just a short part, which is much more challenging. We analyse the 4 task types belonging to long-dependency tasks:

\paragraph{Multiple information retrieval:} This task is quite different from traditional short-term retrieval tasks, there are usually multiple and diverse pieces of evidence throughout the entire text for one specific answer. This task usually involves multi-matching retrieval, but sometimes can also be separable into steps.

\paragraph{Computation:} This task firstly needs multiple information retrieval from a wide range of texts, and then use these data for calculating. A majority of the evidence within the text takes the form of numerical data. However, this task is in fact composed of 3 solvable steps: understand the question to determine which numeric to retrieve, get the numeric through normal retrieval operation (step 1 and 2 may be perform several times to get multiple numeric), calculate the answer based on the retrieved data. Thus it does not belong to logical retrieval, but may sometimes involve multi-matching.

\paragraph{Timeline reorder:} This task requires reordering the timeline of a set of events presented in a permuted order. It apparently needs to compare the size of numbers to determine which event should be retrieved first or later. So it involves logic-based retrieval.

\paragraph{Comprehension and reasoning:} This task demands not only a profound comprehension of the question but also intricate reasoning to discern the underlying implications for searching for the appropriate evidence. It must be a multi-step problem, but whether this issue involves other components remains uncertain and depends on the specific nature of the problem in question.

\subsubsection{Ruler}
We choose 5 difficult tasks from Ruler \citep{hsieh_ruler_2024} to analyse:

\paragraph{Multi-keys NIAH:} Multiple “needles” are inserted into the “haystack”, and only one of them needs to be retrieved. The additional “needles” are hard distractors. This is a normal retrieval task. Though many hard distractors are inserted, powerful LLMs can usually overcome this.

\paragraph{Multi-values NIAH:} Multiple “needles” sharing the same key are inserted into the “haystack”. All values associated with the same key need to be retrieved. This is totally the same as the multi-matching retrieval.

\paragraph{Multi-queries NIAH:} Multiple “needles” with distinct keys are inserted into the “haystack”. On the surface, it may appear that a problem requires the retrieval of multiple values. However, since each value has a distinct key, it can actually be decomposed into multiple simple retrieval tasks. Therefore, this does not fall under the category of logic-based retrieval or multi-match retrieval.

\paragraph{Multi-hop Tracing:} A variable X1 is initialized with a value V, followed by a linear chain of variable name binding statements (e.g., X2 = X1, X3 = X2, ...), which are inserted at various positions of the input. The objective is to return all variable names pointing to the same value V. This is a classic chain-of-retrieval task. It can also be decomposed into multiple simple retrieval tasks, e.g., first retrieve X1, then use X1 as the query to retrieve X2. Therefore, it is multi-step-question, but not multi-matching. It may be logic-based retrieval if the variable name binding statements involve more complex calculations.

\paragraph{Aggregation:} This task includes Common Words (CWE) and Frequent Words Extraction (FWE). A model needs to return the top-K frequent words in the context. This is a very hard multi-step problem consisting of at least 3 steps: identify each word, use each word as the query to do multi-matching retrieval and compare the frequency of each word. So it must involve multi-matching.

\subsubsection{NeedleBench}
NeedleBench \citep{li_needlebench_2024} aims to let NIAH \cite{gkamradt_llmtest_needleinahaystack_2023} more challenging. We analyse all the tasks from it, except the simplest one, Single-Needle Retrieval Task. 

\paragraph{Multi-Needle Retrieval Task:} This task is nearly the same as Multi-queries NIAH. It can actually be decomposed into multiple independent retrieval problems, thus it is not difficult.

\paragraph{Multi-Needle Reasoning:} In this task, the model must first engage in reasoning to comprehend the issue, thereby determining which specific pieces of information are required. Subsequently, it must retrieve these multiple pieces of information from the context. This task is also multi-step-question, which can be solved by CoT \citep{wei_chain--thought_2023}. However, none of the steps necessitates logic-based or multi-matching retrieval.

\paragraph{Ancestral Trace Challenge:} The context encompasses a multitude of interpersonal relationships, and the model is tasked with discerning the ancestral relationship between two individuals. This is similar to Multi-hop Tracing, so it does not necessitate logic-based or multi-matching retrieval.

\subsubsection{Loong}
Loong \citep{wang_leave_2024} is a recent long-context benchmark which emphasized the challenge of the task. Most tasks in it involves mathematical calculation, which is very hard for LLMs. We analyse all of the 4 tasks from it.

\paragraph{Spotlight Locating:} It is aimed at examining the LLMs’ ability to search the evidence within one document from multiple ones. So it is a simple retrieval task.

\paragraph{Comparison:} One of the sub-tasks is that given a specific numerical or conceptual range, the model should output all objects within multiple documents that meet the condition. Apparently, this task is a typical logic-based retrieval task.

\paragraph{Clustering:} One of the sub-tasks requires the model to group the evidence existing in the provided financial reports into corresponding sets based on textual or numerical criteria. Apparently, this task is a typical logic-based retrieval task, too.

\paragraph{Chain of Reasoning:} This task evaluates the model’s proficiency in logical reasoning, which requires LLMs to locate the corresponding evidence within multiple documents and model the logical relationships among them for deducing the answer. This is similar to Multi-hop Tracing, which is a multi-step reasoning task, but whether involving logic-retrieval or multi-matching should depends on specific problems.

\section{Potential Solutions}
\label{app:potential}
As we have found, though CoT can help, relying solely on the LLM itself provide the complete answers with the reasoning process is extremely inefficient. Nevertheless, using external tools may provide a viable solution. For example, if the input context is well-structured (e.g. Markdown tables or JSON data), the model can be prompted to write programs and execute them using external interpreters \citep{chen_program_2023} to accurately yield the correct answer, as demonstrated in Appendix \ref{app:PoT}. However, for more complex and mutable scenarios, it may still require further works such as designing sophisticated systems consisting of multiple AI Agents \citep{xi_rise_2023}. Some recent works \cite{hao_latent_2024} about reasoning in the implicit space may also help.

\end{document}